\documentclass[11pt,a4paper]{article}
\usepackage[hyperref]{naaclhlt2019}
\usepackage{times}
\usepackage{latexsym}
\usepackage{hhline}

\usepackage{multirow}

\usepackage{url}

\usepackage{graphicx}

\usepackage{pgfplotstable}

\usepackage{pgfplots}

\pgfplotsset{compat=1.14}

\aclfinalcopy 

\title{Massively Multilingual Neural Machine Translation}
\author{Roee Aharoni\thanks{~~Work carried out during an internship at Google AI.} \\
        Bar Ilan University \\
        Ramat-Gan \\
        Israel \\
        {\tt roee.aharoni@gmail.com} 
        \And
        Melvin Johnson \and Orhan Firat\\
        Google AI\\
        Mountain View \\
        California\\
        {\tt melvinp,orhanf@google.com} 
        }

\date{}

\begin{document}
\maketitle
\begin{abstract}
  Multilingual neural machine translation (NMT) enables training a single model that supports translation from multiple source languages into multiple target languages. In this paper, we push the limits of multilingual NMT in terms of the number of languages being used. We perform extensive experiments in training massively multilingual NMT models, translating up to 102 languages to and from English within a single model. We explore different setups for training such models and analyze the trade-offs between translation quality and various modeling decisions. We report results on the publicly available TED talks multilingual corpus where we show that massively multilingual many-to-many models are effective in low resource settings, outperforming the previous state-of-the-art while supporting up to 59 languages. Our experiments on a large-scale dataset with 102 languages to and from English and up to one million examples per direction also show promising results, surpassing strong bilingual baselines and encouraging future work on massively multilingual NMT.
\end{abstract}

\section{Introduction}

Neural machine translation (NMT) \cite{kalchbrenner-blunsom:2013:EMNLP,bahdanau2014neural, sutskever2014sequence} is the current state-of-the-art approach for machine translation in both academia \cite{bojar2016findings,bojar2017findings,bojar-EtAl:2018:WMT1} and industry \cite{wu2016google,hassan2018achieving}. Recent works \cite{dong2015multi,firat2016multi,ha2016toward,johnson2017google} extended the approach to support multilingual translation, i.e. training a single model that is capable of translating between multiple language pairs.

Multilingual models are appealing for several reasons. First, they are more efficient in terms of the number of required models and model parameters, enabling simpler deployment. Another benefit is transfer learning; when low-resource language pairs are trained together with high-resource ones, the translation quality may improve \cite{zoph2016transfer,nguyen-chiang:2017:I17-2}. An extreme case of such transfer learning is zero-shot translation \cite{johnson2017google}, where multilingual models are able to translate between language pairs that were never seen during training. 

While very promising, it is still unclear how far one can scale multilingual NMT in terms of the number of languages involved. Previous works on multilingual NMT typically trained models with up to 7 languages \cite{dong2015multi,firat2016zero,ha2016toward,johnson2017google,gu-EtAl:2018:N18-1} and up to 20 trained directions \cite{mauro2017overview} simultaneously. One recent exception is \newcite{neubig2018rapid} who trained many-to-one models from 58 languages into English. While utilizing significantly more languages than previous works, their experiments were restricted to many-to-one models in a low-resource setting with up to 214k examples per language-pair and were evaluated only on four translation directions.

In this work, we take a step towards practical ``universal'' NMT -- training massively multilingual models which support up to 102 languages and with up to one million examples per language-pair simultaneously. Specifically, we focus on training ``English-centric'' many-to-many models, in which the training data is composed of many language pairs that contain English either on the source side or the target side. This is a realistic setting since English parallel data is widely available for many language pairs. We restrict our experiments to Transformer models \cite{vaswani2017attention} as they were shown to be very effective in recent benchmarks \cite{ott2018scaling}, also in the context of multilingual models \cite{lakew2018comparison,sachan2018parameter}.

We evaluate the performance of such massively multilingual models while varying factors like model capacity, the number of trained directions (tasks) and low-resource vs. high-resource settings. Our experiments on the publicly available TED talks dataset \cite{Ye2018WordEmbeddings} show that massively multilingual many-to-many models with up to 58 languages to-and-from English are very effective in low resource settings, allowing to use high-capacity models while avoiding overfitting and achieving superior results to the current state-of-the-art on this dataset \cite{neubig2018rapid,wang2018multilingual} when translating into English.

We then turn to experiment with models trained on 103 languages in a high-resource setting. For this purpose we compile an English-centric in-house dataset, including 102 languages aligned to-and-from English with up to one million examples per language pair. We then train a single model on the resulting 204 translation directions and find that such models outperform strong bilingual baselines by more than 2 BLEU averaged across 10 diverse language pairs, both to-and-from English. Finally, we analyze the trade-offs between the number of involved languages and translation accuracy in such settings, showing that massively multilingual models generalize better to zero-shot scenarios. We hope these results will encourage future research on massively multilingual NMT.


\section{Low-Resource Setting: 59 Languages}

\subsection{Experimental Setup}

The main question we wish to answer in this work is how well a single NMT model can scale to support a very large number of language pairs. The answer is not trivial: on the one hand, training multiple language pairs together may result in transfer learning \cite{zoph2016transfer,nguyen-chiang:2017:I17-2}. This may improve performance as we increase the number of language pairs, since more information can be shared between the different translation tasks, allowing the model to learn which information to share. On the other hand, adding many language pairs may result in a bottleneck; the model has a limited capacity while it needs to handle this large number of translation tasks, and sharing all parameters between the different languages can be sub-optimal \cite{wang-EtAl:2018:EMNLP10} especially if they are not from the same typological language family \cite{sachan2018parameter}. 

We begin tackling this question by experimenting with the TED Talks parallel corpus compiled by \newcite{Ye2018WordEmbeddings}\footnote{\url{github.com/neulab/word-embeddings-for-nmt}}, which is unique in that it includes parallel data from 59 languages. For comparison, this is significantly ``more multilingual'' than the data available from all previous WMT news translation shared task evaluations throughout the years -- the latest being \newcite{bojar2016findings,bojar2017findings,bojar-EtAl:2018:WMT1}, which included 14 languages so far.\footnote{Chinese, Czech, English, Estonian, Finnish, French, German, Hindi, Hungarian, Latvian, Romanian, Russian, Spanish, Turkish. According to \url{http://www.statmt.org/wmtXX}} 

We focus on the setting where we train ``English-centric'' models, i.e. training on all language pairs that contain English in either the source or the target, resulting in 116 translation directions. This dataset is also highly imbalanced, with language pairs including between 3.3k to 214k sentence pairs for training. Table \ref{tab:ted_langs} in the supplementary material details the languages and training set sizes for this dataset. Since the dataset is already tokenized we did not apply additional preprocessing other than applying joint subword segmentation \cite{sennrich2016neural} with 32k symbols.

Regarding the languages we evaluate on, we begin with the same four languages as \newcite{neubig2018rapid} -- Azerbeijani (Az), Belarusian (Be), Galician (Gl) and Slovak (Sk). These languages present an extreme low-resource case, with as few as 4.5k training examples for Belarusian-English. In order to better understand the effect of training set size in these settings, we evaluate on four additional languages that have more than 167k training examples each -- Arabic (Ar), German (De), Hebrew (He) and Italian (It).


\subsection{Model Details}

Using the same data, we trained three massively multilingual models: a many-to-many model which we train using all 116 translation directions with 58 languages to-and-from English, a one-to-many model from English into 58 languages, and a many-to-one model from 58 languages into English. We follow the method of \newcite{ha2016toward,johnson2017google} and add a target-language prefix token to each source sentence to enable many-to-many translation. These different setups enable us to examine the effect of the number of translation tasks on the translation quality as measured in BLEU \cite{papineni-EtAl:2002:ACL}. We also compare our massively multilingual models to bilingual baselines and to two recently published results on this dataset (\newcite{neubig2018rapid,wang2018multilingual}). 

Regarding the models, we focused on the Transformer in the ``Base'' configuration. We refer the reader to \newcite{vaswani2017attention} for more details on the model architecture. 
Specifically, we use 6 layers in both the encoder and the decoder, with model dimension set at 512, hidden dimension size of 2048 and 8 attention heads. We also applied dropout at a rate of 0.2 in the following components: on the sum of the input embeddings and the positional embeddings, on the output of each sub-layer before added to the previous layer input (residual connection), on the inner layer output after the ReLU activation in each feed-forward sub-layer, and to the attention weight in each attention sub-layer. This results in a model with approximately 93M trainable parameters. For all models we used the inverse square root learning rate schedule from \newcite{vaswani2017attention} with learning-rate set at 3 and 40k warmup steps. All models are implemented in Tensorflow-Lingvo \cite{lingvo}.

In all cases we report test results for the checkpoint that performed best on the development set in terms of BLEU. For the multilingual models we create a development set that includes examples we uniformly sample from a concatenation of all the individual language pair development sets, resulting in 13k development examples per model. Another important detail regarding multilingual training is the batching scheme. In all of our multilingual models we use heterogeneous batching, where each batch contains examples which are uniformly sampled from a concatenation of all the language pairs the model is trained on. Specifically, we use batches of 64 examples for sequences shorter than 69 tokens and batches of 16 examples for longer sequences. We did not use over-sampling as the dataset is relatively small.


\subsection{Results}

\begin{table}[!ht]
\begin{center}
\begin{small}
\setlength\tabcolsep{3.8pt}
\begin{tabular}{l|llll|l}
          & Az-En & Be-En & Gl-En & Sk-En & Avg.  \\ \hline
\# of examples  & 5.9k  & 4.5k  & 10k   & 61k   & 20.3k \\ \hline
Neubig \& Hu 18 &       &       &       &               \\       
baselines       & 2.7	& 2.8	& 16.2  & 24    & 11.42 \\
many-to-one     & 11.7  & 18.3  & 29.1  & 28.3  & 21.85 \\ \hline
Wang et al. 18  & 11.82 & 18.71 & 30.3  & 28.77 & 22.4  \\ \hline
Ours  & & & & \\
many-to-one  & 11.24 & 18.28 & 28.63 & 26.78  & 21.23  \\
many-to-many  & \textbf{12.78} & \textbf{21.73} & \textbf{30.65} & \textbf{29.54} & \textbf{23.67} \\ 
\end{tabular}
\caption{X$\rightarrow$En test BLEU on the TED Talks corpus, for the language pairs from \newcite{neubig2018rapid}}
\label{tab:toen_small}
\end{small}
\end{center}
\end{table}


\begin{table}[!ht]
\begin{center}
\begin{small}
\setlength\tabcolsep{4.3pt}
\begin{tabular}{l|llll|l}
          & Ar-En & De-En & He-En & It-En & Avg.  \\ \hline
\# of examples  & 213k  & 167k   & 211k & 203k & 198.5k   \\ \hline
baselines   & 27.84    & 30.5  & \textbf{34.37}   & 33.64  & 31.59   \\
many-to-one  & 25.93 & 28.87 & 30.19 & 32.42 & 29.35  \\
many-to-many  & \textbf{28.32} & \textbf{32.97} & 33.18 & \textbf{35.14} & \textbf{32.4} \\ 
\end{tabular}
\caption{X$\rightarrow$En test BLEU on the TED Talks corpus, for language pairs with more than 167k examples}
\label{tab:toen_larger}
\end{small}
\end{center}
\vspace{-0.3cm}
\end{table}


\begin{figure}[!htp]
\centering
\begin{center}
\includegraphics[width=0.46\textwidth, trim={4.5cm 2cm 4cm 2cm}, clip]{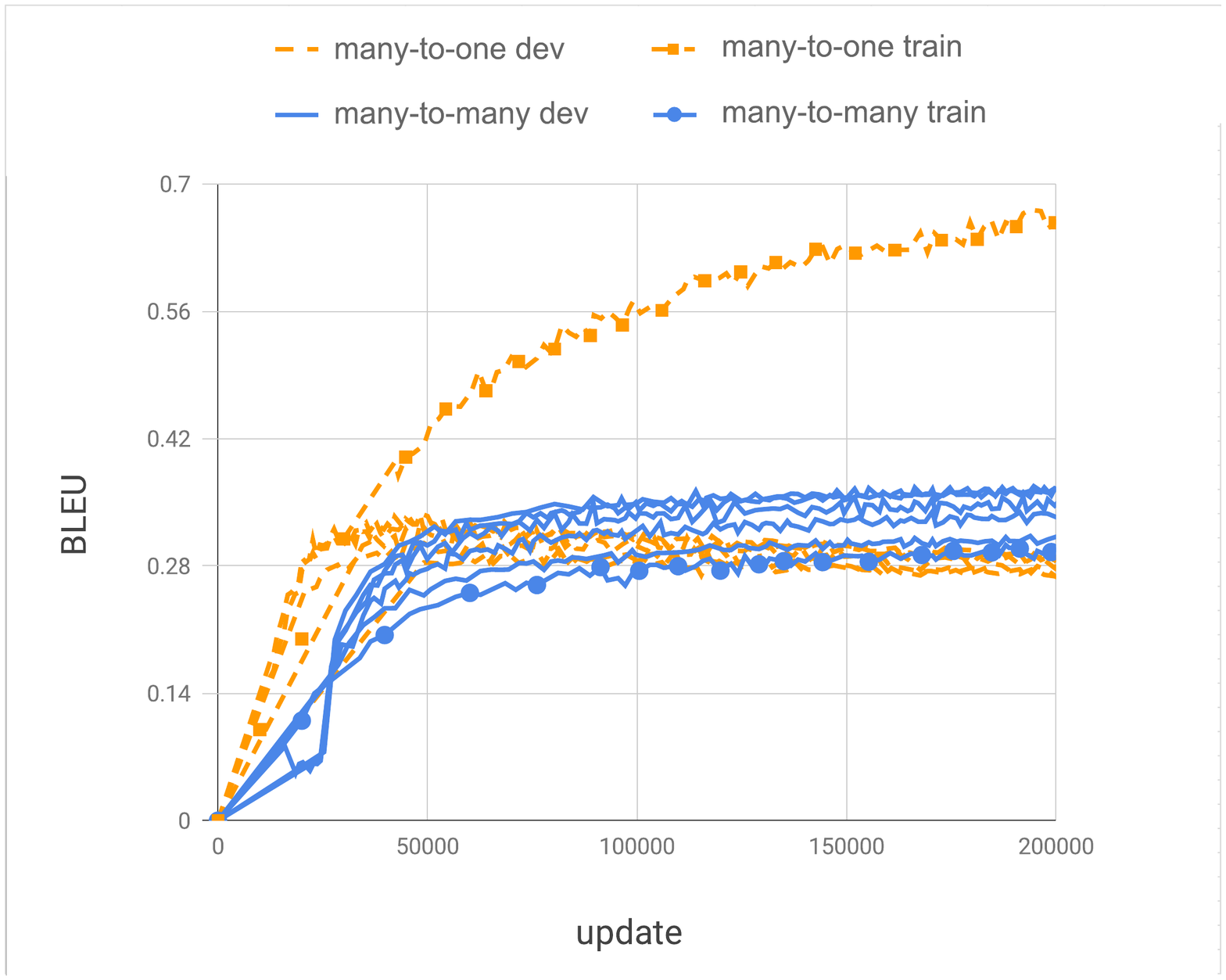}
\caption {Development BLEU on \{It,Ro,Nl,De,Ar\}$\rightarrow$En vs. training BLEU for the many-to-one and many-to-many models. Best viewed in color. \newline}
\label{fig:to_en_dev_train_bleu}
\end{center}
\vspace{-1cm}
\end{figure}

We use tokenized BLEU in order to be comparable with \newcite{neubig2018rapid}. Table \ref{tab:toen_small} shows the results of our experiments when evaluating on the same language pairs as they did. The results under ``Neubig \& Hu 18'' are their bilingual baselines and their best many-to-one models. Their many-to-one models use similar-language-regularization, i.e. fine-tuning a pre-trained many-to-one model with data from the language pair of interest together with data from a language pair that has a typologically-similar source language and more training data (i.e. Russian and Belarusian, Turkish and Azerbaijani). The results under ``Ours'' are our many-to-one and many-to-many models we trained identically in terms of model architecture and hyper-parameters.

We first note that our many-to-many model outperforms all other models when translating into English, with 1.82 BLEU improvement (when averaged across the four language pairs) over the best fine-tuned many-to-one models of \newcite{neubig2018rapid} and 2.44 BLEU improvement over our many-to-one model when averaged across the four low-resource language pairs (Table \ref{tab:toen_small}). This is surprising as it uses the same X$\rightarrow$En data, model architecture and capacity as our many-to-one model, while handling a heavier burden since it also supports 58 \textit{additional} translation tasks (\textit{from} English \textit{into} 58 languages). Our models also outperform the more complex models of \newcite{wang2018multilingual} which use "Soft Decoupled Encoding" for the input tokens, while our models use a simple subword segmentation.

One possible explanation is that the many-to-one model overfits the English side of the corpus as it is multi-way-parallel: in such setting the English sentences are overlapping across the different language pairs, making it much easier for the model to memorize the training set instead of generalizing (when enough capacity is available). On the other hand, the many-to-many model is trained on additional target languages other than English, which can act as regularizers for the X$\rightarrow$En tasks, reducing such overfitting.

To further illustrate this, Figure \ref{fig:to_en_dev_train_bleu} tracks the BLEU scores on the individual development sets during training for Italian (It), Romanian (Ro), Dutch (Nl), German (De) and Arabic (Ar) into English (left), together with BLEU scores on a subset of the training set for each model. We can see that while the many-to-one model degrades in performance on the development set, the many-to-many model still improves. Note the large gap in the many-to-one model between the training set BLEU and the development set BLEU, which points on the generalization issue that is not present in the many-to-many setting.
We also note that our many-to-one model is on average 0.75 BLEU behind the best many-to-one models in \newcite{neubig2018rapid}. We attribute this to the fact that their models are fine-tuned using similar-language-regularization while our model is not.


We find an additional difference between the results on the resource-scarce languages (Table \ref{tab:toen_small}) and the higher-resource languages (Table \ref{tab:toen_larger}). Specifically, the bilingual baselines outperform the many-to-one models only in the higher-resource setting. This makes sense as in the low-resource setting the baselines have very few training examples to outperform the many-to-one models, while in the higher resource setting they have access to more training data. This corroborates the results of \newcite{gu-EtAl:2018:N18-1} that showed the sensitivity of such models to similar low resource conditions and the improvements gained from using many-to-one models (however with much fewer language pairs).


\begin{table}[t]
\begin{center}
\begin{small}
\setlength\tabcolsep{4.1pt}
\begin{tabular}{l|llll|l}
             & En-Az & En-Be & En-Gl & En-Sk & Avg. \\ \hline
\# of examples     & 5.9k  & 4.5k  & 10k   & 61k   & 20.3k  \\ \hline
baselines    & 2.16  & 2.47  & 3.26  & 5.8   & 3.42  \\
one-to-many  & \textbf{5.06} & \textbf{10.72} & \textbf{26.59} & \textbf{24.52} & \textbf{16.72} \\
many-to-many & 3.9	 & 7.24	 & 23.78 & 21.83 & 14.19 \\
\multicolumn{0}{l}{}\\ 

              & En-Ar & En-De & En-He & En-It & Avg.  \\ \hline
\# of examples      & 213k  & 167k   & 211k & 203k & 198.5k   \\ \hline
baselines     & 12.95  & 23.31   & 23.66  & 30.33  & 22.56   \\
one-to-many   & \textbf{16.67} & \textbf{30.54} & \textbf{27.62} & \textbf{35.89} & \textbf{27.68}   \\
many-to-many  & 14.25 & 27.95 & 24.16 & 33.26 & 24.9   \\ 
\end{tabular}
\caption{En$\rightarrow$X test BLEU on the TED Talks corpus}
\label{tab:ento}
\end{small}
\end{center}
\vspace{-0.6cm}
\end{table}

Table \ref{tab:ento} shows the results of our massively multilingual models and bilingual baselines when evaluated out-of-English. In this case we see an opposite trend: the many-to-many model performs worse than the one-to-many model by 2.53 BLEU on average. While previous works \cite{wang-EtAl:2018:EMNLP10, sachan2018parameter} discuss the phenomena of quality degradation in English-to-many settings, this shows that increasing the number of \textit{source} languages also causes additional degradation in a many-to-many model. This degradation may be due to the English-centric setting: since most of the translation directions the model is trained on are into English, this leaves less capacity for the other target languages (while still performing better than the bilingual baselines on all 8 language pairs). We also note that in this case the results are consistent among the higher and lower resource pairs -- the one-to-many model is better than the many-to-many model, which outperforms the bilingual baselines in all cases. This is unlike the difference we saw in the X$\rightarrow{}$En experiments since here we do not have the multi-way-parallel overfitting issue.

\subsection{Discussion}
From the above experiments we learn that NMT models can scale to 59 languages in a low-resource, imbalanced, English-centric setting, with the following observations: (1) massively multilingual many-to-many models outperform many-to-one and bilingual models with similar capacity and identical training conditions when averaged over 8 language pairs into English. We attribute this improvement over the many-to-one models to the multiple target language pairs which may act as regularizers, especially in this low-resource multi-way-parallel setting that is prone to memorization. (2) many-to-many models are inferior in performance when going out-of-English in comparison to a one-to-many model. We attribute this to English being over-represented in the English-centric many-to-many setting, where it appears as a target language in 58 out of 116 trained directions, which may harm the performance on the rest of the target languages as the model capacity is limited.\footnote{This issue may be alleviated by over-sampling the non-English-target pairs, but we leave this for future work.}

It is important to stress the fact that we compared the different models under \textit{identical training conditions} and did not perform extensive hyper-parameter tuning for each setting separately. However, we believe that such tuning may improve performance even further, as the diversity in each training batch is very different between the different settings. For example, while the baseline model batches include only one language in the source and one language in the target, the many-to-many model includes 59 languages in each side with a strong bias towards English. These differences may require tailored hyper-parameter choices for each settings (i.e. different batch sizes, learning rate schedules, dropout rates etc.) which would be interesting to explore in future work. 

In the following experiments we investigate whether these observations hold using (1) an even larger set of languages, and (2) a much larger, balanced training corpus that is not multi-way-parallel.

\section{High-Resource Setting: 103 Languages}

\subsection{Experimental Setup}
In this setting we scale the number of languages and examples per language pair further when training a single massively multilingual model. Since we are not aware of a publicly available resource for this purpose, we construct an in-house dataset. This dataset includes 102 language pairs which we ``mirror'' to-and-from English, with up to one million examples per language pair. This results in 103 languages in total, and 204 translation directions which we train simultaneously. More details about this dataset are available in Table \ref{tab:hundred_stats}, and Table \ref{tab:in_house_langs} in the supplementary material details all the languages in the dataset.\footnote{The average number of examples per language pair is 940k, as for 13 out of the 102 pairs we had less than one million examples available.}

\begin{table}[!b]
\begin{center}
\begin{tabular}{|l|l|}\hline
\# of language pairs             & 102         \\\hline
examples per pair             &             \\
\hspace{1cm}min               & 63,879      \\
\hspace{1cm}max               & 1,000,000   \\
\hspace{1cm}average           & 940,087     \\
\hspace{1cm}std. deviation    & 188,194     \\\hline
total \# of examples & 95,888,938 \\\hline
\end{tabular}
\end{center}
\caption{Training set details for the 103 langauges corpus, X$\rightarrow$En data.}
\label{tab:hundred_stats}
\end{table}

\begin{table*}[!ht]
\begin{center}
\setlength\tabcolsep{4.9pt}
\begin{tabular}{l|llllllllll|l}
           & Ar    & Az    & Be    & De     & He    & It    & Nl    & Ro    & Sk    & Tr    & Avg.   \\ \hline
baselines    & 23.34 & 16.3 & 21.93 & 30.18  & 31.83 & \textbf{36.47}  & 36.12 & 34.59 & 25.39 & 27.13 & 28.33 \\
many-to-one  & \textbf{26.04} & \textbf{23.68} & \textbf{25.36} & 35.05 & \textbf{33.61} & 35.69 & \textbf{36.28} & 36.33 & 28.35 & \textbf{29.75}  & \textbf{31.01} \\ 
many-to-many & 22.17 & 21.45 & 23.03 & \textbf{37.06} & 30.71 & 35.0 & 36.18 & \textbf{36.57} & \textbf{29.87} & 27.64 & 29.97 \\
\end{tabular}
\end{center}
\caption{X$\rightarrow$En test BLEU on the 103-language corpus}
\label{tab:toen100}
\end{table*}

\begin{table*}[!ht]
\begin{center}
\setlength\tabcolsep{4.9pt}
\begin{tabular}{l|llllllllll|l}
           & Ar    & Az    & Be    & De   & He    & It    & Nl    & Ro    & Sk    & Tr    & Avg.  \\ \hline
baselines    & 10.57 & 8.07  & 15.3 & 23.24 & 19.47 & 31.42 & 28.68 & 27.92 & 11.08 & 15.54 & 19.13 \\
one-to-many  & \textbf{12.08} & \textbf{9.92}  & \textbf{15.6} & \textbf{31.39}  & \textbf{20.01} & \textbf{33} & \textbf{31.06} & \textbf{28.43} & \textbf{17.67} & \textbf{17.68} & \textbf{21.68}     \\ 
many-to-many & 10.57  & 9.84  & 14.3 & 28.48 & 17.91 & 30.39 & 29.67 & 26.23 & 18.15 & 15.58 & 20.11 
\end{tabular}
\end{center}
\caption{En$\rightarrow$X test BLEU on the 103-language corpus}
\label{tab:ento100}
\end{table*}

Similarly to our previous experiments, we compare the massively multilingual models to bilingual baselines trained on the same data. We tokenize the data using an in-house tokenizer and then apply joint subword segmentation to achieve an open-vocabulary. In this setting we used a vocabulary of 64k subwords rather than 32k. Since the dataset contains 24k unique characters, a 32k symbol vocabulary will consist of mostly characters, thereby increasing the average sequence length. Regarding the model, for these experiments we use a larger Transformer model with 6 layers in both the encoder and the decoder, model dimension set to 1024, hidden dimension size of 8192, and 16 attention heads. This results in a model with approximately 473.7M parameters.\footnote{This is larger than the Transformer ``Big'' configuration, which includes approximately 213M trained parameters.} Since the model and data are much larger in this case, we used a dropout rate of 0.1 for our multilingual models and tuned it to 0.3 for our baseline models as it improved the translation quality on the development set.

We evaluate our models on 10 languages from different typological families: \textit{Semitic} -- Arabic (Ar), Hebrew (He), \textit{Romance} -- Galician (Gl), Italian (It), Romanian (Ro), \textit{Germanic} -- German (De), Dutch (Nl), \textit{Slavic} -- Belarusian (Be), Slovak (Sk) and \textit{Turkic} -- Azerbaijani (Az) and Turkish (Tr). We evaluate both to-and-from English, where each language pair is trained on up to one million examples. As in the previous experiment, we report test results from the model that performed best in terms of BLEU on the development set. 

\subsection{Results}
Table \ref{tab:toen100} describes the results when translating into English. First, we can see that both multilingual models perform better than the baselines in terms of average BLEU. This shows that massively multilingual many-to-many models can work well in realistic settings with millions of training examples, 102 languages and 204 jointly trained directions to-and-from English. Looking more closely, we note several different behaviors in comparison to the low-resource experiments on the TED Talks corpus. First, the many-to-one model here performs better than the many-to-many model. This shows that the previous result was indeed due to the pathologies of the low-resource dataset; when the training data is large enough and not multi-way-parallel there is no overfitting in the many-to-one model, and it outperforms the many-to-many model in most cases while they are trained identically. 

One particular outlier in this case is German-to-English, where the many-to-one model is 2 BLEU points below the many-to-many model. We examine the BLEU score of this language pair on its dedicated German-English development set during training in the many-to-one model and find that it highly fluctuates. We then measure the performance on the test set for this language pair by choosing the best checkpoint on the dedicated German-English development set (instead of on the mixed multilingual development set) and find it to be 38.07, which is actually \textit{higher} in 1 BLEU than the best result of the many-to-many model. This shows that while training many languages together, there is no ``silver bullet'': some languages may suffer from severe interference during training (i.e. a reduction of 3 BLEU in this case, from 38.07 to 35.05) while other languages continue to improve with more updates.

Table \ref{tab:ento100} describes the results when translating out-of-English. Again, both of the massively multilingual models perform better than the baselines when averaged across the 10 evaluated language pairs, while handling up to 102 languages to-and-from English and 204 translation tasks simultaneously. In this case the results are similar to those we observed on the TED talks corpus, where the one-to-many model performs better than the many-to-many model. Again, this advantage may be due to the one-to-many model handling a smaller number of tasks while not being biased towards English in the target side like the many-to-many model.

\section{Analysis}

\begin{table*}[!ht]
\begin{center}
\begin{tabular}{l|llllllll|l}
           & Ar-En & En-Ar & Fr-En & En-Fr & Ru-En & En-Ru & Uk-En & En-Uk & Avg.  \\ \hline
5-to-5     & \textbf{23.87} & \textbf{12.42} & \textbf{38.99} & \textbf{37.3} & 29.07 & \textbf{24.86} & \textbf{26.17} & 16.48 & \textbf{26.14} \\
25-to-25   & 23.43 & 11.77 & 38.87 & 36.79 & \textbf{29.36} & 23.24 & 25.81 & \textbf{17.17} & 25.8 \\
50-to-50   & 23.7  & 11.65 & 37.81 & 35.83 & 29.22 & 21.95 & 26.02 & 15.32  & 25.18 \\
75-to-75   & 22.23 & 10.69  & 37.97 & 34.35 & 28.55 & 20.7 & 25.89 & 14.59 & 24.37 \\
103-to-103 & 21.16 & 10.25  & 35.91 & 34.42 & 27.25 & 19.9 & 24.53 & 13.89 & 23.41 \\
\end{tabular}
\caption{Supervised performance while varying the number of languages involved}
\label{tab:supervised}
\end{center}
\vspace{-0.6cm}
\end{table*}

\begin{table}[!ht]
\begin{center}
\begin{small}
\begin{tabular}{l|llll|l}
           & Ar-Fr & Fr-Ar & Ru-Uk & Uk-Ru & Avg.  \\ \hline
5-to-5     & 1.66  & 4.49  & 3.7  & 3.02  & 3.21  \\
25-to-25   & 1.83  & \textbf{5.52}  & \textbf{16.67} & 4.31  & 7.08   \\
50-to-50   & \textbf{4.34} & 4.72  & 15.14 & \textbf{20.23} & \textbf{11.1} \\
75-to-75   & 1.85  & 4.26  & 11.2  & 15.88 & 8.3  \\
103-to-103 & 2.87  & 3.05  & 12.3  & 18.49 & 9.17  \\
\end{tabular}
\caption{Zero-Shot performance while varying the number of languages involved}
\label{tab:zero-shot}
\end{small}
\end{center}
\vspace{-0.6cm}
\end{table}

The above results show that massively multilingual NMT is indeed possible in large scale settings and can improve performance over strong bilingual baselines. However, it was shown in a somewhat extreme case with more than 100 languages trained jointly, where we saw that in some cases the joint training may harm the performance for some language pairs (i.e. German-English above). In the following analysis we would like to better understand the trade-off between the number of languages involved and the translation accuracy while keeping the model capacity and training configuration fixed.

\subsection{Multilinguality \& Supervised Performance}
We first study the effect of varying the number of languages on the translation accuracy in a supervised setting, where we focus on many-to-many models. We create four subsets of the in-house dataset by sub-sampling it to a different number of languages in each subset. In this way we create four additional English-centric datasets, containing 5, 25, 50 and 75 languages each to-and-from English. We make sure that each subset contains all the languages from the next smaller subsets -- i.e. the 25 language subset contains the 5 language subset, the 50 language subset contains the 25 language subset and so on. We train a similar-capacity large Transformer model (with 473.7M parameters) on each of these subsets and measure the performance for each model on the 8 supervised language pairs from the smallest subset -- \{Arabic, French, Russian, Ukrainian\}$\leftrightarrow$English. In this way we can analyze to what extent adding more languages improves or harms translation quality while keeping the model capacity fixed, testing the capacity vs. accuracy ``saturation point''.


Table \ref{tab:supervised} shows the results of this experiment, reporting the test results for the models that performed best on the multilingual development set. We can see that in most cases the best results are obtained using the 5-to-5 model, showing that there is indeed a trade off between the number of languages and translation accuracy when using a fixed model capacity and the same training setup. One may expect that the gaps between the different models should become smaller and even close with more updates, as the models with more languages see less examples per language in each batch, thus requiring more updates to improve in terms of BLEU. However, in our setting these gaps did not close even after the models converged, leaving 2.73 average BLEU difference between the 5-to-5 and the 103-to-103 model. 

\begin{figure}[!b]
\centering
\includegraphics[width=0.475\textwidth, trim={2.8cm 2.2cm 4.4cm 2.2cm},clip]{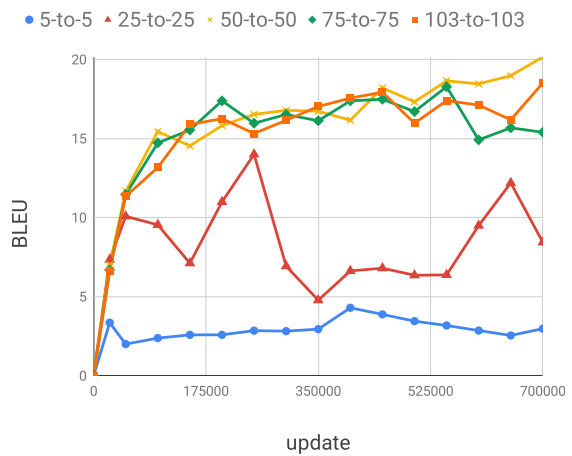}
\vspace{-0.2cm}
\caption {Zero-shot BLEU during training for Ukranian to Russian}
\label{fig:zero_shot}
\vspace{-0.5cm}
\end{figure}

\subsection{Multilinguality \& Zero-Shot Performance}

We then study the effect of the number of languages on zero-shot translation accuracy. Since we find zero-shot accuracy as an interesting measure for model generalization, we hypothesize that by adding more languages, the model is forced to create a more generalized representation to better utilize its capacity, which may improve zero-shot performance. We choose four language pairs for this purpose: Arabic$\leftrightarrow$French which are distant languages, and Ukrainian$\leftrightarrow$Russian which are similar. 
Table \ref{tab:zero-shot} shows the results of our models on these language pairs. For Arabic$\leftrightarrow$French the BLEU scores are very low in all cases, with the 50-to-50 and 25-to-25 models being slightly better than rest on Ar-Fr and Fr-Ar respectively. On Russian$\leftrightarrow$Ukrainian we see clear improvements when increasing the number of languages to more than five. 

Figure \ref{fig:zero_shot} further illustrates this, showing the better generalization performance of the massively multilingual models under this zero-shot setting. While the zero-shot performance in this case is low and unstable for the 5-to-5 and 25-to-25 models, it is much better for the 50-to-50, 75-to-75 and 103-to-103 models. Given these results we can say that the balance between capacity and generalization here favors the mid range 50-to-50 model, even when using models with more than 473M trained parameters. This may hint at the necessity of even larger models for such settings, which is a challenging avenue for future work. We also note that our 103 language corpus includes up to one million examples per language pair -- while in real-world MT deployments, systems are trained on much more examples per pair. This again emphasizes the need for better techniques for training such massively multilingual models as we may already be hitting the capacity barrier in our setting.


\section{Related Work}

\newcite{dong2015multi} extended the NMT model of \newcite{bahdanau2014neural} to one-to-many translation (from English into 4 languages) by adding a dedicated decoder per target language, showing improvements over strong single-pair baselines. \newcite{firat2016multi,firat2016zero} proposed many-to-many models (with up to 6 languages) by using separate encoders and decoders per language while sharing the attention mechanism. They also introduced the notion of zero-resource translation, where they use synthetic training data generated through pivoting to train translation directions without available training data. \newcite{ha2016toward} and \newcite{johnson2017google} proposed to use a shared encoder-decoder-attention model for many-to-many translation (with up to 7 languages in the latter). In order to determine the target language in such scenarios they proposed adding dedicated target-language symbols to the source. This method enabled zero-shot translation, showing the ability of the model to generalize to unseen pairs.

Recent works propose different methods for parameter sharing between language pairs in multilingual NMT. \newcite{blackwood2018multilingual} propose sharing all parameters but the attention mechanism and show improvements over sharing all parameters. \newcite{sachan2018parameter} explore sharing various components in self-attentional (Transformer) models. \newcite{lu2018neural} add a shared ``interlingua'' layer while using separate encoders and decoders. \newcite{zaremoodi-buntine-haffari:2018:Short} utilize recurrent units with multiple blocks together with a trainable routing network. \newcite{platanios2018contextual} propose to share the entire network, while using a contextual parameter generator that learns to generate the parameters of the system given the desired source and target languages. \newcite{gu-EtAl:2018:N18-1} propose a ``Universal Language Representation'' layer together with a Mixture-of-Language-Experts component to improve a many-to-one model from 5 languages into English.

While the mentioned studies provide valuable contributions to improving multilingual models, they apply their models on only up to 7 languages \cite{johnson2017google} and 20 trained directions \cite{mauro2017overview} in a single model, whereas we focus on scaling NMT to much larger numbers of languages and trained directions. Regarding massively multilingual models, \newcite{neubig2018rapid} explored methods for rapid adaptation of NMT to new languages by training multilingual models on the 59-language TED Talks corpus and fine-tuning them using data from the new languages. While modeling significantly more languages than previous studies, they only train many-to-one models, which we show are inferior in comparison to our proposed massively multilingual many-to-many models when evaluated into English on this dataset. 

\newcite{tiedemann2018emerging} trained an English-centric many-to-many model on translations of the bible including 927 languages. While this work pointed to an interesting phenomena in the latent space learned by the model where it clusters representations of typologically-similar languages together, it did not include any evaluation of the produced translations. Similarly, \newcite{malaviya18} trained a many-to-English system including 1017 languages from bible translations, and used it to infer typological features for the different languages (without evaluating the translation quality). In another relevant work, \newcite{artetxe2018massively} trained an NMT model on 93 languages and used the learned representations to perform cross-lingual transfer learning. Again, they did not report the performance of the translation model learned in that massively multilingual setting.

\section{Conclusions and Future Work}
We showed that NMT models can successfully scale to 102 languages to-and-from English with 204 trained directions and up to one million examples per direction. Such models improve the translation quality over similar single-pair baselines when evaluated to and from English by more than 2 BLEU when averaged over 10 diverse language pairs in each case. We show a similar result on the low-resource TED Talks corpus with 59 languages and 116 trained directions. We analyze the trade-offs between translation quality and the number of languages involved, pointing on capacity bottlenecks even with very large models and showing that massively multilingual models can generalize better to zero-shot settings.

We hope this work will encourage future research on massively multilingual NMT, enabling easier support for systems that can serve more people around the globe. There are many possible avenues for future work, including semi-supervised learning in such settings, exploring ways to reduce the performance degradation when increasing the number of languages, or using such models for multilingual transfer learning \cite{mccann2017learned,eriguchi2018zero,artetxe2018massively}. Understanding and improving zero-shot performance in such scenarios is also a promising direction for future work.

\section*{Acknowledgments}
We would like to thank the Google Brain and Google Translate teams for their useful inputs and discussions. We would also like to thank the entire Lingvo development team for their foundational contributions to this project. \\

\bibliography{main}
\bibliographystyle{acl_natbib}

\appendix


\section{Supplementary Material}
\begin{table}[!b]
\begin{small}
\begin{center}
\begin{tabular}{l|l}
Language         & Train set size \\ \hline
Arabic           & 214111         \\
Hebrew           & 211819         \\
Russian          & 208458         \\
Korean           & 205640         \\
Italian          & 204503         \\
Japanese         & 204090         \\
Chinese-Taiwan   & 202646         \\
Chinese-China    & 199855         \\
Spanish          & 196026         \\
French           & 192304         \\
Portuguese-Brazil& 184755         \\
Dutch            & 183767         \\
Turkish          & 182470         \\
Romanian         & 180484         \\
Polish           & 176169         \\
Bulgarian        & 174444         \\
Vietnamese       & 171995         \\
German           & 167888         \\
Persian          & 150965         \\
Hungarian        & 147219         \\
Serbian          & 136898         \\
Greek            & 134327         \\
Croatian         & 122091         \\
Ukrainian        & 108495         \\
Czech            & 103093         \\
Thai             & 98064          \\
Indonesian       & 87406          \\
Slovak           & 61470          \\
Swedish          & 56647          \\
Portuguese       & 51785          \\
Danish           & 44940          \\
Albanian         & 44525          \\
Lithuanian       & 41919          \\
Macedonian       & 25335          \\
Finnish          & 24222          \\
Burmese          & 21497          \\
Armenian         & 21360          \\
French-Canadian  & 19870          \\
Slovenian        & 19831          \\
Hindi            & 18798          \\
Norwegian        & 15825          \\
Georgian         & 13193          \\
Estonian         & 10738          \\
Kurdish          & 10371          \\
Galician         & 10017          \\
Marathi          & 9840           \\
Mongolian        & 7607           \\
Esperanto        & 6535           \\
Tamil            & 6224           \\
Urdu             & 5977           \\
Azerbaijani      & 5946           \\
Bosnian          & 5664           \\
Chinese          & 5534           \\
Malay            & 5220           \\
Basque           & 5182           \\
Bengali          & 4649           \\
Belarusian       & 4509           \\
Kazakh           & 3317          
\end{tabular}
\end{center}
\end{small}
\caption {Language pairs in the TED talks dataset (58 languages, paired with English) with the train-set size for each pair.}
\label{tab:ted_langs}
\end{table}

\begin{table}[!b]
\centering
\begin{small}
\begin{tabular}{ll}
\multicolumn{2}{l}{Languages}  \\ \hline
Afrikaans      & Laothian      \\
Albanian       & Latin         \\
Amharic        & Latvian       \\
Arabic         & Lithuanian    \\
Armenian       & Luxembourgish*\\
Azerbaijani    & Macedonian    \\
Basque         & Malagasy      \\
Belarusian     & Malay         \\
Bengali        & Malayalam     \\
Bosnian        & Maltese       \\
Bulgarian      & Maori         \\
Burmese        & Marathi       \\
Catalan        & Mongolian     \\
Cebuano        & Nepali        \\
Chichewa*      & Norwegian     \\
Chinese        & Pashto        \\
Corsican*      & Persian       \\
Croatian       & Polish        \\
Czech          & Portuguese    \\
Danish         & Punjabi       \\
Dutch          & Romanian      \\
Esperanto      & Russian       \\
Estonian       & Samoan*       \\
Finnish        & Scots Gaelic* \\
French         & Serbian       \\
Frisian        & Sesotho       \\
Galician       & Shona*        \\
Georgian       & Sindhi*       \\
German         & Sinhalese     \\
Greek          & Slovak        \\
Gujarati       & Slovenian     \\
Haitian Creole & Somali        \\
Hausa*         & Spanish       \\
Hawaiian*      & Sundanese     \\
Hebrew         & Swahili       \\
Hindi          & Swedish       \\
Hmong*         & Tagalog       \\
Hungarian      & Tajik*        \\
Icelandic      & Tamil         \\
Igbo           & Telugu        \\
Indonesian     & Thai          \\
Irish          & Turkish       \\
Italian        & Ukrainian     \\
Japanese       & Urdu          \\
Javanese       & Uzbek         \\
Kannada        & Vietnamese    \\
Kazakh         & Welsh         \\
Khmer          & Xhosa         \\
Korean         & Yiddish       \\
Kurdish        & Yoruba*       \\
Kyrgyz         & Zulu         
\end{tabular}
\end{small}
\caption{Language pairs in the in-house dataset (102 languages, paired with English). For languages marked with * we had less than 1M examples, while for the rest we used exactly 1M.}
\label{tab:in_house_langs}
\end{table}

\end{document}